\documentclass[letterpaper]{article} 
\usepackage{aaai2026}  
\usepackage{times}  
\usepackage{helvet}  
\usepackage{courier}  
\usepackage[hyphens]{url}  
\usepackage{graphicx} 
\usepackage{natbib}  
\usepackage{caption} 
\usepackage{booktabs}
\usepackage{multirow}
\usepackage{array}
\usepackage{xcolor}
\usepackage{colortbl}
\usepackage{adjustbox}
\usepackage{xcolor}
\usepackage[most]{tcolorbox}
\usepackage[table]{xcolor} 
\usepackage{hhline} 
\usepackage{url}

\usepackage{algorithm}
\usepackage{algorithmic}
\usepackage{threeparttable}
\usepackage{booktabs}          
\usepackage{multirow}          
\usepackage{threeparttable}    
\usepackage{graphicx}          
\usepackage{xcolor}            
\usepackage{colortbl}          
\usepackage{array}             
\usepackage{arydshln}          
\usepackage[utf8]{inputenc} 

\definecolor{specblue}{RGB}{190, 220, 245}

\newtcolorbox{quotebox}{
    colback=specblue!60, 
    colframe=gray!30,
    coltext=black,
    boxrule=0pt,
    arc=4pt,
    left=5pt,
    right=5pt,
    top=5pt,
    bottom=5pt,
}
\usepackage[utf8]{inputenc}
\usepackage{xcolor}
\usepackage{tcolorbox}
\usepackage{graphicx}
\usepackage{lipsum} 
\usepackage{float}
\usepackage{tcolorbox}
\definecolor{lightblue}{RGB}{173,216,230}
\definecolor{darkblue}{RGB}{70,130,180}

\tcbset{
    mybox/.style={
        colback=lightblue,
        colframe=darkblue,
        boxrule=2pt,
        arc=4pt,
        left=10pt,
        right=10pt,
        top=10pt,
        bottom=10pt,
        fonttitle=\bfseries,
        title={Text and Figure Integration Example}
    }
}

\usepackage{newfloat}
\usepackage{listings}
\DeclareCaptionStyle{ruled}{labelfont=normalfont,labelsep=colon,strut=off} 
\floatstyle{ruled}
\newfloat{listing}{tb}{lst}{}
\floatname{listing}{Listing}
\newcommand{\ourbench}{\textit{ThinkWithImages}-\textsc{PRMBench}}

\title{What, Whether and How? Unveiling Process Reward Models \\for \textit{Thinking with Images} Reasoning}
\author{
    Yujin Zhou\textsuperscript{\rm 1}\equalcontrib,
    Pengcheng Wen\textsuperscript{\rm 1}\equalcontrib,
    Jiale Chen\textsuperscript{\rm 2}\equalcontrib,
    Boqin Yin\textsuperscript{\rm 1},
    Han Zhu\textsuperscript{\rm 1},
    Jiaming Ji\textsuperscript{\rm 3},
    Juntao Dai\textsuperscript{\rm 3},
    Chi-Min Chan\textsuperscript{\rm 1},
    Sirui Han\textsuperscript{\rm 1}\thanks{Corresponding author.}
}
\affiliations{
    \textsuperscript{\rm 1}Hong Kong University of Science and Technology \\
    \textsuperscript{\rm 2}Sun Yat-sen University \\
    \textsuperscript{\rm 3}Peking University \\

\{yzhouha,pc.wen\}@connect.ust.hk~~ertiam047@gmail.com~~siruihan@ust.hk
}

\usepackage{bibentry}

\begin{document}

\maketitle

\begin{abstract}

The rapid advancement of Large Vision Language Models (LVLMs) has demonstrated excellent abilities in various visual tasks. Building upon these developments, the \textit{thinking with images} paradigm has emerged, enabling models to dynamically edit and re-encode visual information at each reasoning step, mirroring human visual processing. However, this paradigm introduces significant challenges as diverse errors may occur during reasoning processes. This necessitates Process Reward Models (PRMs) for distinguishing positive and negative reasoning steps, yet existing benchmarks for PRMs are predominantly text-centric and lack comprehensive assessment under this paradigm. To address these gaps, this work introduces the first comprehensive benchmark specifically designed for evaluating PRMs under the \textit{thinking with images} paradigm. Our main contributions are: (1) Through extensive analysis of reasoning trajectories and guided search experiments with PRMs, we define 7 fine-grained error types and demonstrate both the necessity for specialized PRMs and the potential for improvement. (2) We construct a comprehensive benchmark comprising 1,206 manually annotated thinking with images reasoning trajectories spanning 4 categories and 16 subcategories for fine-grained evaluation of PRMs. (3) Our experimental analysis reveals that current LVLMs fall short as effective PRMs, exhibiting limited capabilities in visual reasoning process evaluation with significant performance disparities across error types, positive evaluation bias, and sensitivity to reasoning step positions. These findings demonstrate the effectiveness of our benchmark and establish crucial foundations for advancing PRMs in LVLMs.

\end{abstract}

\section{Introduction}

The human visual system is dynamically heterogeneous in scale , freely switching between broad views and detailed focus as needed~\citep{wandell1995foundations,wang2025lsnet}. In contrast, conventional Large Vision Language Models (LVLMs) are fundamentally limited by their reliance on static tokenization of image information for language model comprehension—a process that introduces inevitable information loss and constrains the exploitation of fine-grained visual details~\citep{bordes2024introduction,ghosh2024exploring}. Recent breakthroughs in LVLMs, exemplified by OpenAI's o3~\citep{Thinkingwithimages}, have introduced the transformative \textit{thinking with images} paradigm~\citep{su2025thinking} to overcome these challenges. Unlike traditional approaches that isolate visual perception from reasoning, \textit{thinking with images} paradigm mirrors the dynamic nature of human visual processing by enabling models to proactively edit and re-encode visual information at each reasoning step using integrated image processing tools, thereby mitigating static tokenization limitations and enhancing overall visual understanding~\citep{zheng2025deepeyes,sarch2025grounded,su2025pixel,hu2024visual,sun2024visual,cheng2025visual,zhang2025chain,fan2025grit,zhu2025safemtmultiturnsafetymultimodal}.

\begin{figure}[!t]
    \centering
    \includegraphics[width=1.0\linewidth]{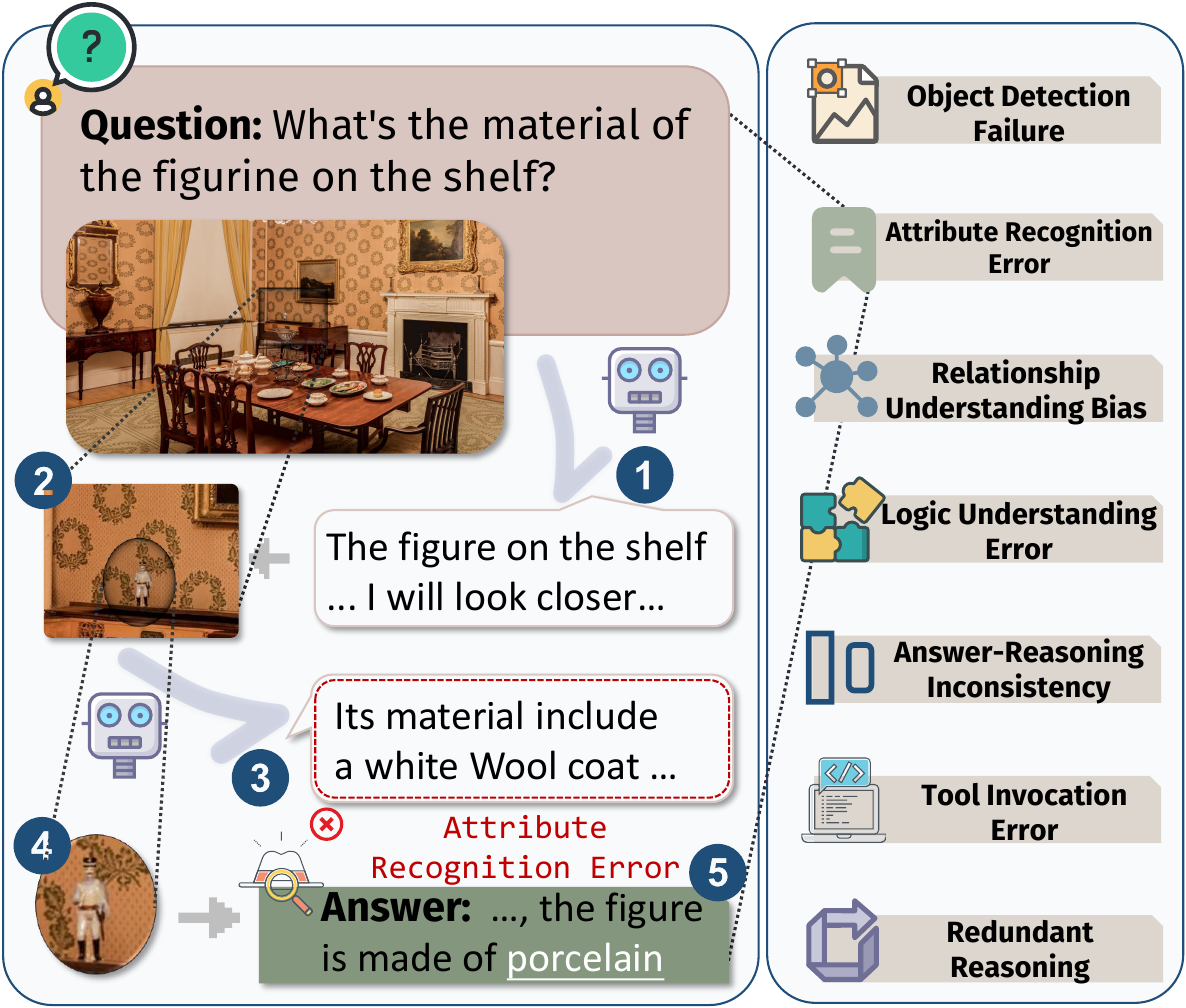}
    \caption{\textbf{Identified Error Types in \textit{Thinking with Images} Paradigm.} We identify and categorize seven distinct error types from reasoning trajectories of current \textit{thinking with images} models. Some are inherent LVLMs limitations and some are novel errors introduced by this paradigm.}
    \label{fig:main_figure}
\end{figure}

However, despite these significant advances, LVLMs under \textit{thinking with images} paradigm still encounter several remaining issues that hinder their reasoning capabilities, as illustrated in Figure~\ref{fig:main_figure}. 

Through comprehensive manual analysis of 7,558 reasoning trajectories generated by four \textit{thinking with images} models across four established benchmarks, we systematically identified and taxonomized seven distinct categories of remaining reasoning errors under the \textit{thinking with images} paradigm.
These error categories highlight ongoing challenges that stem from both the inherent limitations of traditional LVLM and new issues introduced by the dynamic nature of the \textit{thinking with images} paradigm, ultimately compromising the quality of the overall reasoning.

Inspired by the remarkable success of Process Reward Models (PRMs) in enhancing verbal reasoning through step-wise supervision~\citep{chan2025j1,chan2025boosting,wang2025visualprm,lightman2023let,wen2025thinkpatterns,cao2025towards,shi-etal-2025-legalreasoner,zhang2024rest,cao2025safelawbench}, we observe that reasoning trajectories within the \textit{thinking with images} paradigm can be naturally decomposed into discrete interleaved text-image steps. This structural compatibility naturally motivates us to investigate a meaningful question for the \textit{thinking with images} paradigm:
\begin{center}
\textbf{\textit{Whether Process Reward Models Help?}}
\end{center}
In the absence of specialized PRMs for \textit{thinking with images} paradigm, we explored prompting existing state-of-the-art LVLMs to function as PRMs for guided search evaluation across five benchmarks. Our preliminary experiments demonstrate that LVLMs can improve reasoning performance over baseline approaches. 

To further analyze the abilities of existing models to identify errors within the paradigm of \textit{ thinking with images} and to facilitate future development of more effective specialized PRMs, \textbf{we introduce~\ourbench, a comprehensive and fine-grained benchmark designed to assess PRMs under the paradigm of \textit{thinking with images}}. Unlike existing process-level benchmarks that focus primarily on text-based trajectories, \ourbench~addresses the unique challenges of evaluating PRMs in \textit{thinking with images} scenarios. Our benchmark comprises 1,206 meticulously curated instances spanning 4 major categories and 16 subcategories, with quality validated by five expert annotators. We implement controlled curation methodologies to maintain consistent difficulty levels across visual reasoning domains, while covering diverse scenarios including geometric analysis, spatial relationships, temporal dynamics, and multi-object interactions. 

Equipped with~\ourbench, we conducted extensive evaluations across over 10 state-of-the-art models. Through quantitative analysis and qualitative observations, our key contributions are summarized as follows:

\begin{itemize}
    \item \textbf{Fine-grained error type taxonomy}: Through extensive experiments and observations, \textbf{7} fine-grained error types are defined under \textit{thinking with images} paradigm, systematically capturing and categorizing common visual reasoning failure modes.
    \item \textbf{Necessity and potential for improvements of PRMs}: Guided search experiments demonstrate that existing LVLMs can help when serving as PRMs but show significant room for improvement.
    \item \textbf{The first \textit{thinking with images} PRM benchmark}: A curated collection of \textbf{1,206} manually annotated high-quality \textit{thinking with images} reasoning trajectories spanning \textbf{4} categories and \textbf{16} subcategories, enabling fine-grained evaluation of existing LVLMs as PRMs.
    \item \textbf{Comprehensive experimental analysis and in-depth insights}: Reveals current LVLMs fall short as PRMs, demonstrating limited capability in capturing vision reasoning process evaluation with significant performance disparities across different error types. Analysis uncovers consistent positive bias in evaluations and notable step sensitivity with significant performance variations during multi-step reasoning processes.
\end{itemize}

\section{Related Works}

\paragraph{Large Vision Language Model Reasoning with Tools}
The integration of reasoning capabilities with visual understanding, enabling models to reason through interleaved text and image sequences, has emerged as a crucial frontier in multimodal reasoning known as \textit{thinking with images} paradigm~\citep{su2025thinking, Thinkingwithimages}. Current approaches for enhancing visual reasoning in LVLMs can be categorized into two directions: (1) Training-free methods: Early works explored prompt-based tool integration and other training-free approaches, enabling models to leverage external visual processing tools through carefully designed prompts, instructions, or predefined pipelines~\citep{hu2024visual,shen2024zoomeye,wang2025retrieval,li2025dyfo}. (2) Training-based methods: A second line of work focuses on training models to acquire visual reasoning capabilities through data-driven approaches. This includes supervised fine-tuning with specialized datasets to enable models to invoke image processing tools during reasoning processes, as well as reinforcement learning approaches that employ reward-based optimization, typically computing rewards based on final answer correctness to incentivize models to follow \textit{thinking with images} paradigm~\citep{zheng2025deepeyes,su2025pixel,cao2025ground,sarch2025grounded,jiang2025vlm,wu2025reinforcing,zhang2025chain,su2025openthinkimg,liu2025visual}. Our work focuses primarily on models obtained through training-based methods, analyzing their reasoning trajectories, and utilizing them as upstream \textit{thinking with images} models for systematic investigation.

\begin{figure*}[!t]
    \centering
    \includegraphics[width=1.0\linewidth]{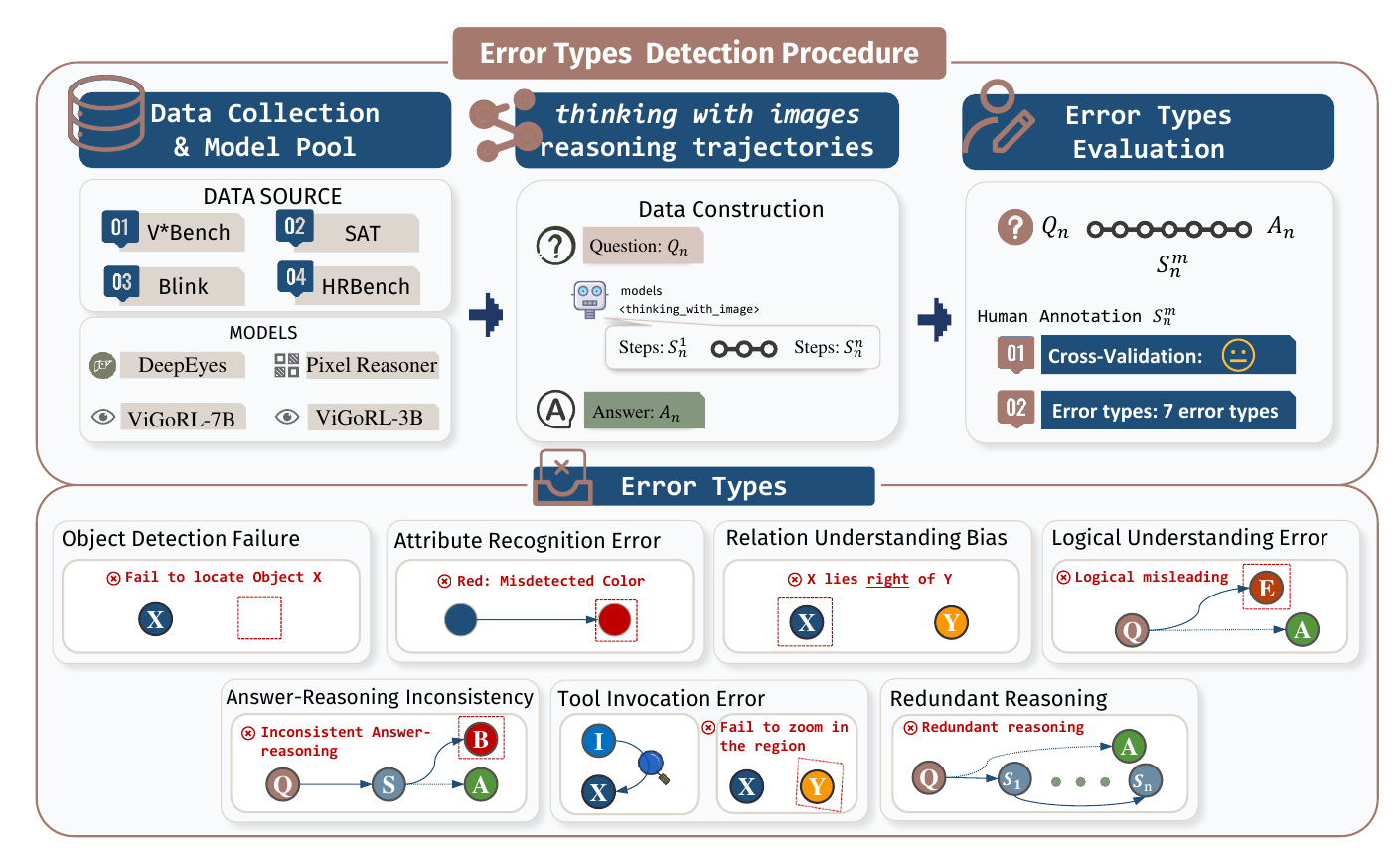}
    \caption{\textbf{Error Types Detection Procedure for \textit{Thinking with Images} Paradigm.} We collect extensive reasoning trajectories by deploying four \textit{ thinking with images} models across four benchmarks. Through systematic analysis and categorization of these trajectories, we identify seven distinct errors that commonly occur in \textit{ thinking with images} paradigm.}
    \label{fig:error_type}
\end{figure*}

\paragraph{Benchmarks for Process Reward Models}

As PRMs have gained increasing attention for their ability to provide step wise supervision during reasoning, effectively and comprehensively evaluating their capabilities has become crucial~\citep{luo2024improve,feng2025prm}. This has led to the development of several PRM benchmarks, including PRMBench~\citep{song2025prmbench}, VisualPRM~\citep{wang2025visualprm}, MPBench~\citep{xu2025mpbench}, and ProcessBench~\citep{zheng2024processbench}, which evaluate PRMs' ability to identify fine-grained errors in upstream models' reasoning trajectories. However, existing benchmarks primarily focus on evaluating PRMs for text-only reasoning trajectories. We introduce~\ourbench, the first benchmark specifically designed to evaluate PRMs' effectiveness in detecting errors within interleaved text-image reasoning trajectories under the \textit{thinking with images} paradigm.

\section{Study Setup}

We begin by constructing a comprehensive collection of \textit{thinking with images} reasoning trajectories, which serves as the foundation for our systematic investigation of reasoning errors, PRM effectiveness evaluation, and~\ourbench development. Using four representative \textit{thinking with images} models across five benchmarks, we collect 7,558 reasoning trajectories.

\subsection{\textit{Thinking with Images} LVLMs}

We select four state-of-the-art models that demonstrate representative capabilities in multimodal reasoning with integrated visual processing tools: ViGoRL-3B and 7B~\citep{sarch2025grounded}, DeepEyes-7B~\citep{zheng2025deepeyes}, and PixelReasoner-7B~\citep{su2025pixel}. These models are chosen based on their ability to perform dynamic visual processing operations such as cropping, zooming,
which are fundamental characteristics of \textit{thinking with images} paradigm.

\subsection{Downstream Benchmarks}

We evaluate our selected models across five established benchmarks that cover diverse visual reasoning scenarios: V*Bench~\citep{huang2024vbench}, Blink~\citep{fu2024blink}, SAT-2~\citep{ray2024sat}, HRBench~\citep{wang2025divide}, and MME-RealWorld~\citep{zhang2024mme}. These benchmarks collectively provide comprehensive coverage of visual reasoning tasks including geometric analysis, spatial relationships, temporal dynamics, and complex multi-object interactions, ensuring our analysis captures the full spectrum of challenges encountered in \textit{thinking with images} applications.

\section{What Issues Persist in Current\\ \textit{Thinking with Images} LVLMs?}\label{sec:remain_issues}


Through extensive experiments and observation of large-scale trajectories, we identify and categorize seven distinct error types that serve as both our annotation guidelines and incorrectness classification criteria:

\begin{figure*}[!t]
    \centering
    \includegraphics[width=1.0\linewidth]{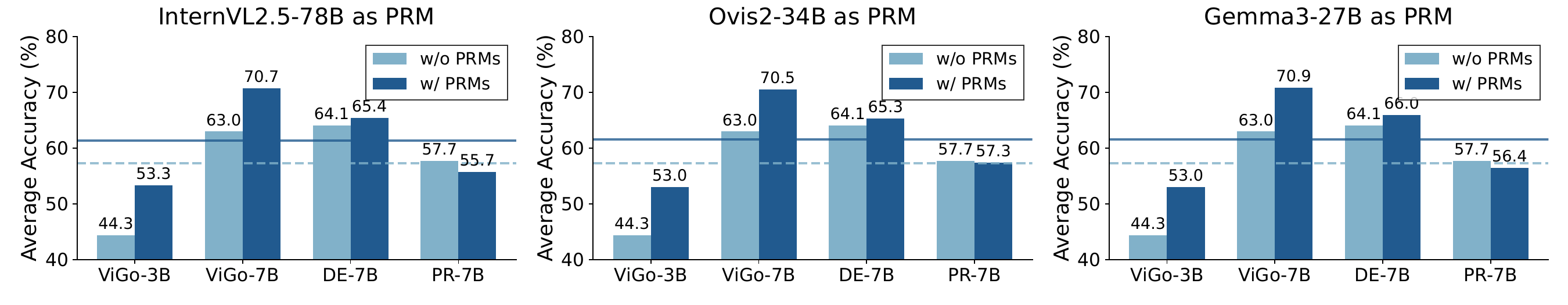}
    \caption{\textbf{Performance Comparison of \textit{Thinking with Images} Models with and without LVLMs as PRMs.} Results demonstrate a general trend toward improved performance when using LVLMs as PRMs, though improvements vary across models.}
    \label{fig:prms_results}
\end{figure*}

\noindent\textbf{Object Detection Failure (ODF):} The model fails to locate or detect objects within the image, resulting in missed visual elements critical to the reasoning process.

\noindent\textbf{Attribute Recognition Error (AE):} While the model successfully localizes objects, it fails to correctly identify their attributes. This includes confusion regarding colors (red vs. orange), materials (metal vs. plastic), shapes, or states (open vs. closed).

\noindent\textbf{Relationship Understanding Bias (RU):} Inaccurate or ambiguous comprehension of spatial relationships (``left'', ``above'', ``between''), action relationships (``holding'', ``chasing''), or comparative relationships (``larger'', ``closer'').

\noindent\textbf{Logical Understanding Error (LU):} The model demonstrates an inability to comprehend the fundamental question or task requirements, leading to irrelevant or illogical reasoning paths.

\noindent\textbf{Answer-Reasoning Inconsistency (ARI):} In the final step, the model produces a correct answer that does not align with the preceding reasoning process, often characterized by hasty conclusion-drawing without proper justification.

\noindent\textbf{Tool Invocation Error (TE):} Failure to correctly utilize available tools for object localization or visual analysis, resulting in inadequate visual processing.

\noindent\textbf{Redundant Reasoning (RR):} Situations where the answer is determined in an early step (e.g., step 1), making subsequent chain-of-thought steps superfluous and unnecessary for the final conclusion.

\section{Whether Process Reward Models Can Help?}
\label{sec:prm_help}
To investigate whether existing LVLMs can effectively serve as PRMs for \textit{thinking with images} paradigm, we conduct preliminary experiments using various state-of-the-art models as surrogate PRMs.
\begin{algorithm}[t!]
\caption{PRM Guided Search for \textit{Thinking with Images}}
\label{alg:guided}
\begin{algorithmic}[1]
    \STATE \textbf{Input:} Problem $p$, LVLM $\mathcal{M}$, PRM $\mathcal{R}$, Beam width $k$, Max steps $T$
    \STATE \textbf{Output:} Best interleaved trajectory $\tau^*$
    \STATE $\mathcal{B} \gets \{(\text{init\_state}, \emptyset, 0)\}$ \COMMENT{Initialize beam}
    \STATE $\tau^* \gets \text{None}$, $\text{best\_score} \gets -\infty$

    \FOR{$t = 1$ \TO $T$}
        \STATE $\mathcal{C} \gets \emptyset$ 
        \FOR{each $(s, \tau, \text{score})$ in $\mathcal{B}$}
            \STATE $\text{steps} \gets \mathcal{M}.\text{Generate}(s, k)$ \COMMENT{Generate text-image interleaved steps}
            \FOR{each step $a$ in $\text{steps}$}
                \STATE $s' \gets \text{Apply}(s, a)$, $\tau' \gets \tau \cup \{a\}$ \COMMENT{$a$ contains text and image}
                \STATE $r \gets \mathcal{R}.\text{Score}(\tau')$ \COMMENT{PRM evaluates interleaved trajectory}
                \STATE Add $(s', \tau', \text{score} + r)$ to $\mathcal{C}$
            \ENDFOR
        \ENDFOR
        \STATE $\mathcal{B} \gets \text{TopK}(\mathcal{C}, k)$ \COMMENT{Keep top-k candidates}
    \ENDFOR
    \STATE \textbf{return} Best interleaved trajectory from $\mathcal{B}$
\end{algorithmic}
\label{alg:prm_guided_search}
\end{algorithm}

\subsection{Experimental Setup}

\noindent\textbf{Model Selection.} We select three representative LVLMs spanning different model families and sizes to serve as surrogate PRMs: Gemma3-27B, Ovis2-34B, and InternVL2.5-78B. These models analyze reasoning trajectories generated under the \textit{thinking with images} paradigm, where each trajectory consists of interleaved text-image steps with visual information processed throughout the workflow.

\noindent\textbf{Evaluation Protocol.} We employ \textbf{Guided Search} as our primary evaluation framework to assess PRM effectiveness. In this approach, candidate PRMs evaluate intermediate reasoning steps and provide real-time feedback to guide trajectory generation, as outlined in Algorithm~\ref{alg:guided}.

\noindent\textbf{Guided Search Implementation.} For upstream models, we generate $k=8$ step-by-step responses at each reasoning stage. PRMs score each candidate response, and the highest-scored response is selected (with ties broken by order). The process continues under a $\text{maxstep}=10$ constraint.

\noindent\textbf{Evaluation Metrics.} We measure performance using accuracy as our primary evaluation metric, and demonstrate PRM effectiveness by comparing LVLM performance before and after PRM integration.

\subsection{Performance Analysis}

\label{sec:prm_help}

\begin{figure*}[ht]
    \centering
    \includegraphics[width=1.0\linewidth]{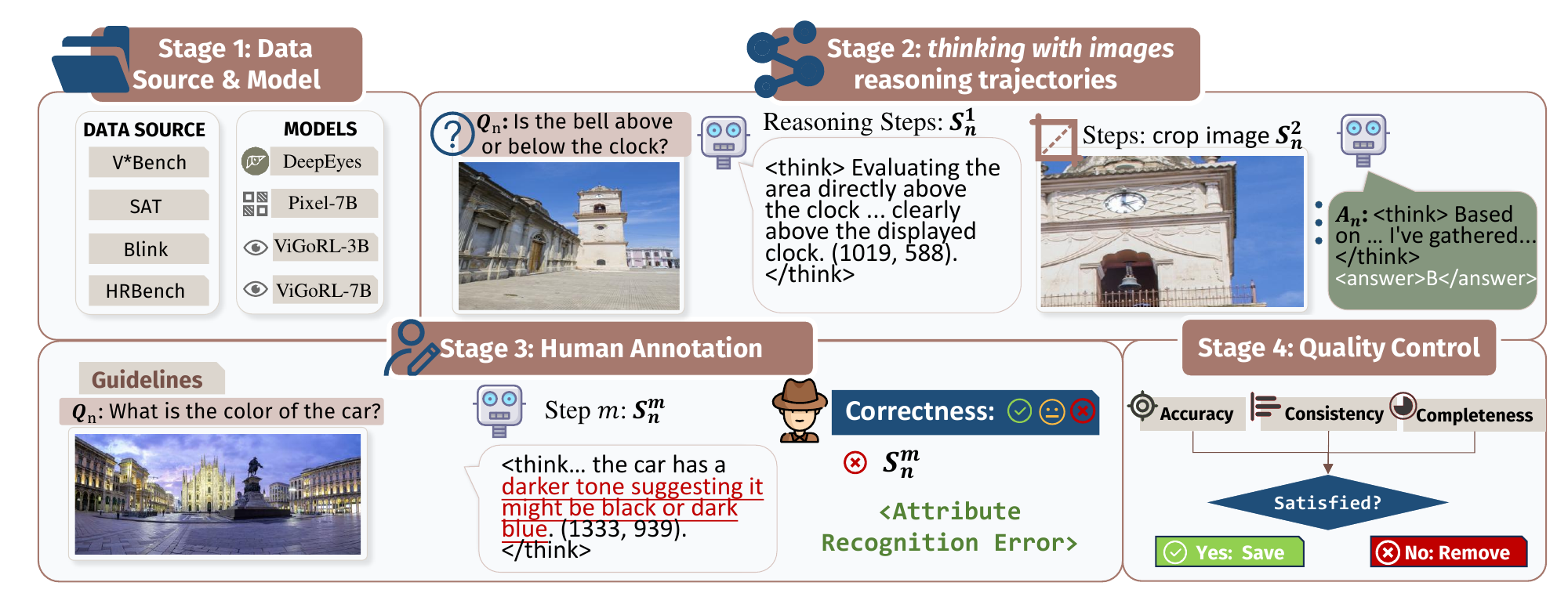}
    \caption{\textbf{Construction Pipeline for \ourbench.} Building upon the reasoning trajectories collected as described in Figure \ref{fig:error_type}, we perform step-by-step manual annotation for each trajectory and classify erroneous steps into seven error types. A comprehensive quality control process filters trajectories based on three key criteria: accuracy, consistency, and completeness, resulting in the curated \ourbench.}
    \label{fig:pipeline}
\end{figure*}

Figure~\ref{fig:prms_results} demonstrates the overall effectiveness of LVLMs as PRMs in advancing \textit{thinking with images} paradigm. While PixelReasoner-7B shows minimal to no improvement with PRMs, other models demonstrate consistent gains ranging from 1-8\%, with PRMs generally outperforming non-PRM approaches across all evaluated models. These findings suggest that PRMs are effective under \textit{thinking with images} paradigm, and there is still room for further improvement.

\section{How Can We Build Better PRMs?\\ Introducing \ourbench}
\label{sec:bench}
To build more effective PRMs, we introduce \ourbench~to support further advancements. In this section, we outline its construction pipeline and assess the performance of existing LVLMs as PRMs using our benchmark, aiming to highlight current challenges.

\subsection{\ourbench~Construction Pipeline}

\ourbench~is a meticulously designed multi-modal process reward benchmark tailored for \textit{thinking with images} paradigm. Our construction comprises four key stages that ensure the scalability, accuracy, and reliability.

\paragraph{Stage 1: Data Collection} We systematically collect question-image pairs from diverse open-source datasets that align with \textit{thinking with images} paradigm. Our data collection strategy encompasses four primary sources: VBench, HRBench, SAT, and BLINK. These datasets are strategically selected to ensure comprehensive coverage across diverse visual reasoning scenarios and task complexities.

\paragraph{Stage 2: Trajectory Construction} For each question-image pair collected from the open-source datasets, we employ our diverse model pool to generate multiple reasoning trajectories. This stage produces multiple reasoning trajectories for each query, providing a rich foundation for subsequent annotation.

\paragraph{Stage 3: Human Annotation} The annotation process consists of two critical phases designed to ensure data quality and consistency:

\noindent\textbf{Data Filtering:} From the various reasoning trajectories generated in Stage 2, we first apply rule-based methods to deduplicate each query, retaining only one valid trajectory per question to ensure quality and uniqueness.

\noindent\textbf{Manual Annotation:} We utilize the open-source Label Studio platform with carefully designed annotation guidelines. Five annotators work over seven days to annotate each step based on our established guidelines and incorrect classification criteria. We filter out low-quality reasoning chains to maintain benchmark integrity.

\paragraph{Stage 4: Quality Assurance}

For the reasoning trajectories retained from Stage 3, three annotators conduct quality verification over three days in batches. Each question is evaluated across three dimensions: accuracy, consistency, and completeness. Only trajectories that satisfied all three criteria are retained. Through this rigorous quality assurance process, we finally collect 1,206 valid reasoning trajectories that make up our final \ourbench.

\begin{figure}[t]
    \centering
    \includegraphics[width=1.0\linewidth]{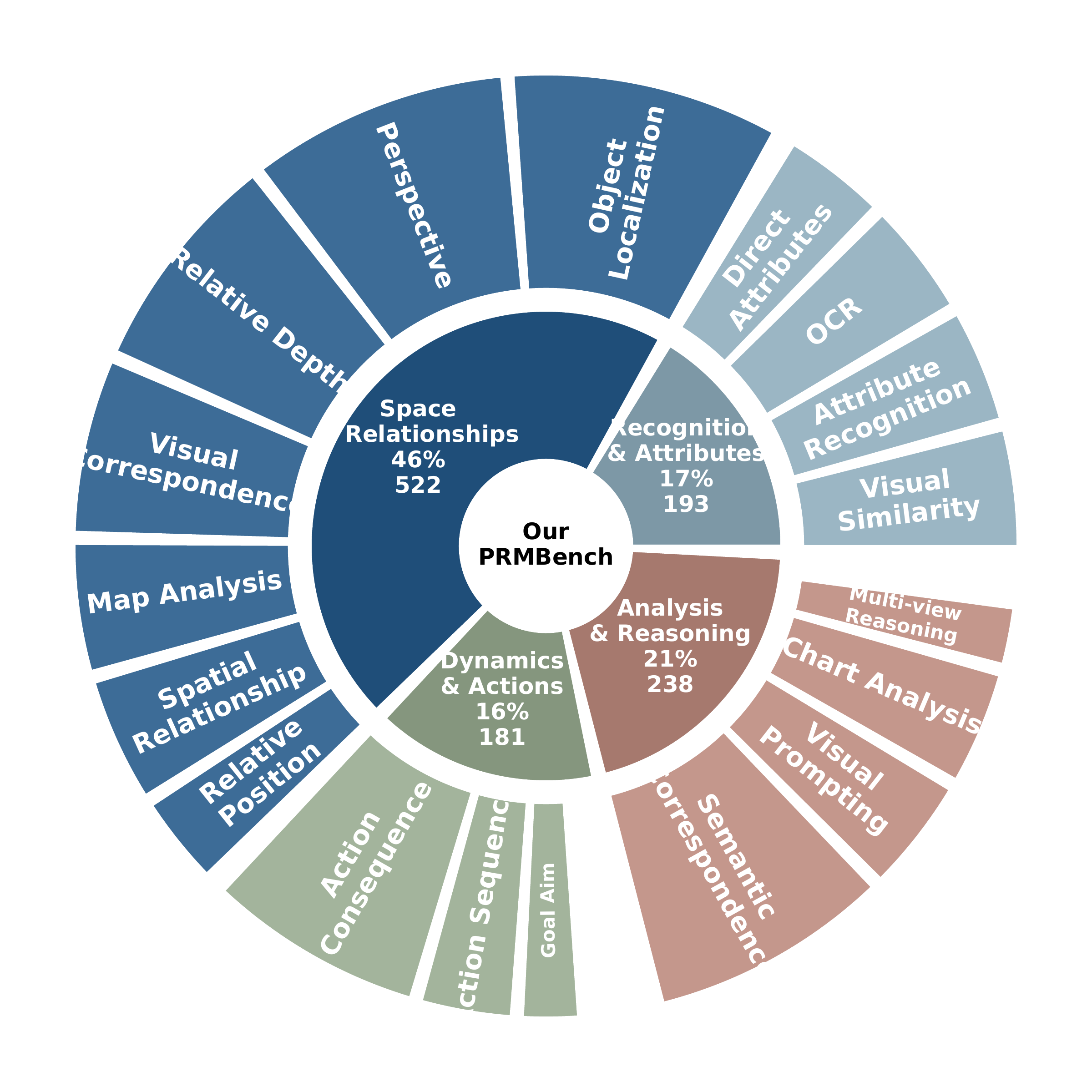}
    \caption{Composition of~\ourbench.}
    \label{fig:bubble}
\end{figure}

\begin{figure}[t]
    \centering
    \includegraphics[width=0.85\linewidth]{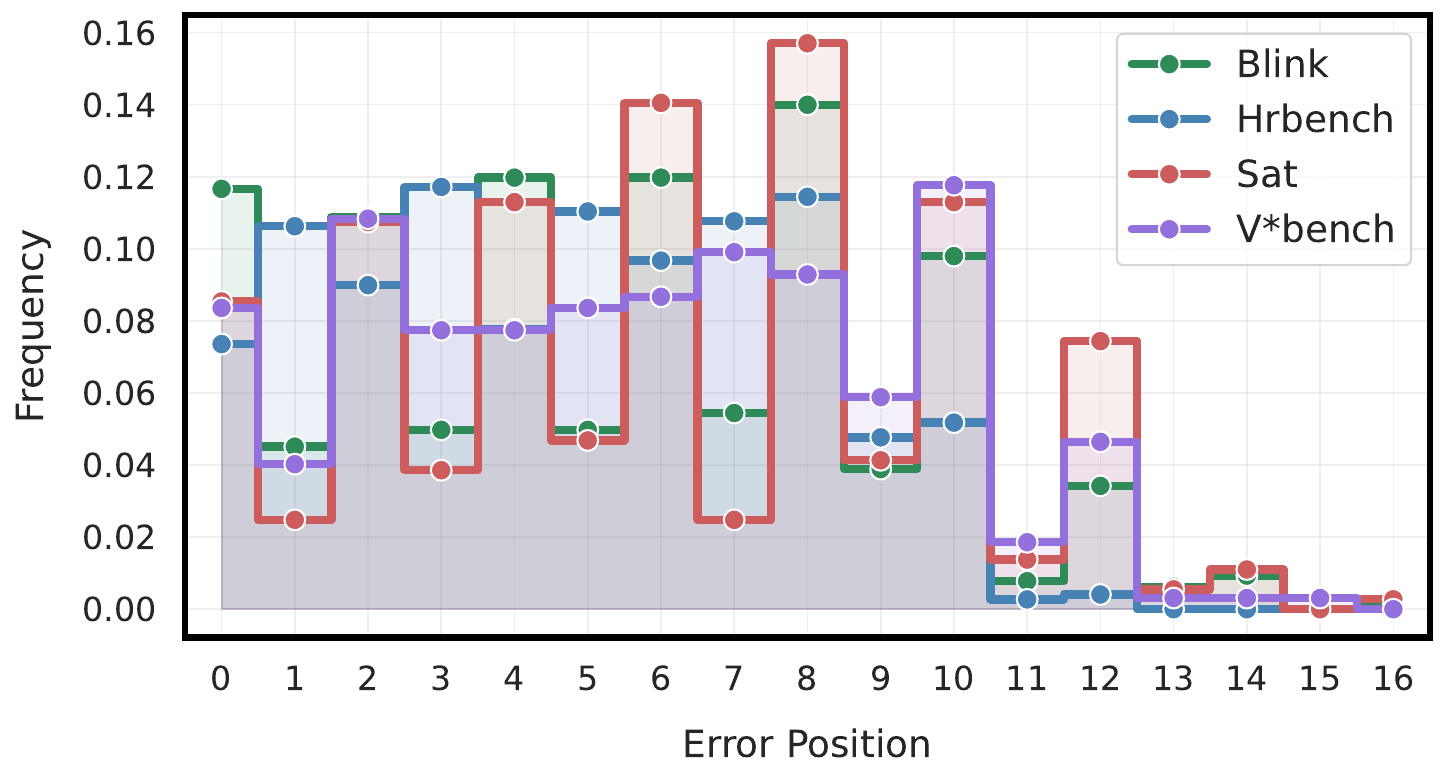}
    \caption{Distribution of Error Steps.}
    \label{fig:error_position}
\end{figure}

This comprehensive four-stage pipeline ensures that our proposed bench provides a robust foundation for evaluating multi-modal PRMs in visual reasoning tasks. Our resulting benchmark systematically covers four fundamental dimensions: Recognition \& Attributes, Space \& Relationships, Dynamics \& Actions, and Analysis \& Reasoning, encompassing 16 specific categories of visual reasoning tasks as illustrated in Figure~\ref{fig:bubble}. The final dataset comprises 1,206 questions with corresponding image reasoning chains, with comprehensive statistics presented in Table~\ref{tab:dataset-stats}. Figure~\ref{fig:error_position} shows the distribution of error positions across reasoning steps.

\begin{table}[t]
    \centering
    \label{tab:dataset-stats}
    \begin{tabular}{l|cc}
    \toprule
    \textbf{Benchmark Statistics}  & \textbf{Count} & \textbf{Percentage} \\
    \midrule
    \textbf{Total Samples} & \boxed{1,206} & 100.0\% \\
    • V*bench & 134 & 11.2\% \\
    • Blink & 478 & 39.7\% \\
    • Sat & 284 & 23.4\% \\
    • HRbench & 310 & 25.7\% \\
    \midrule
    \textbf{Total Steps} & \boxed{12,714} & 100.0\% \\
    • Correct Steps & 6,714 & 52.8\% \\
    • Incorrect Steps & 5,233 & 41.2\% \\
    • Neutral Steps & 767 & 6.0\% \\
    \midrule
    \small\textbf{\texttt{Total Images} }& \boxed{6,544} & -- \\
    \small\textbf{\texttt{Average Steps}}& \boxed{10.41} & -- \\
    \bottomrule
    \end{tabular}
    \caption{Statistics of~\ourbench.}
    \label{tab:dataset-stats}
\end{table}

\section{Evaluation}

\subsection{Experimental Setup}
\subsubsection{Models}
To comprehensively evaluate LVLMs as PRMs on~\ourbench, we test open-source models including LLaVA-OneVision (7B, 72B)~\citep{li2024llava}, Qwen2.5-VL (7B, 72B)~\citep{bai2025qwen2}, InternVL2.5 (8B, 78B)~\citep{chen2024expanding,zhu2025internvl3}, Gemma3 (4B, 27B)~\citep{team2025gemma}, and Ovis2 (8B, 34B)~\citep{lu2024ovis}, alongside proprietary models GPT-4o~\citep{achiam2023gpt} and Gemini-2.5-Flash~\citep{comanici2025gemini}. Human evaluation results are in Table~\ref{tab:main_results}.

\subsubsection{Evaluation Metrics}

We evaluate models using accuracy (ACC) and F1 scores. Accuracy measures the proportion of correctly classified steps. F1 mitigates class imbalance between correct and incorrect steps.

\subsection{Main Results}

\begin{table*}[ht]
\centering
\begin{tabular}{l@{\hspace{16 pt}}c@{\hspace{12pt}}c@{\hspace{12pt}}c@{\hspace{12pt}}c@{\hspace{12pt}}c@{\hspace{12pt}}c@{\hspace{12pt}}c@{\hspace{12pt}}c@{\hspace{12pt}}c} 
    \toprule
    \textbf{Models$\downarrow$ Metrics$\rightarrow$}  & \textbf{ACC} & \textbf{F1} & \texttt{TE} & \texttt{RU} & \texttt{AE} & \texttt{RR} & \texttt{ODF} & \texttt{ARI} & \texttt{LU}\\
    \midrule
    \multicolumn{10}{c}{\textit{Proprietary Large Vision Language Models}} \\
    \midrule
    GPT-4o & 31.45 & 41.34 & 4.61 & 40.10 & 52.94 & 33.33 & 64.02 & 79.31 & 58.21 \\
    Gemini-2.5-Flash & 36.81 & 43.56 & 38.21 & 44.93 & 61.76 & 70.83 &  61.51 & 51.72 & 49.25 \\
    \midrule
    \multicolumn{10}{c}{\textit{Open-Source Large Vision Language Models}} \\
    \midrule
    Qwen2.5-VL-7B & 24.89 & 8.84 & 1.78 & 5.34 & 7.27 & 6.33 & 10.50 & 10.00 & 3.48 \\
    Qwen2.5-VL-72B & 28.27 & 32.81 & 20.36 & 24.93 & 37.27 & 26.58 & 37.9 & 41.67 & 27.83 \\
    \midrule
    InternVL2.5-8B & 22.77 & 13.30 & 2.55 & 8.4 & 7.79 & 1.41 & 18.18 & 20.45 & 10.94 \\
    InternVL2.5-78B & 21.13 & 24.28 & 0.32 & 17.2 & 31.17 & 5.63 & 30.00 & 43.18 & 21.88 \\
    \midrule
    LLaVA-OneVision-7B & 52.36 & 2.55 & 0.20 & 1.78 & 1.01 & 1.27 & 3.65 & 0.80 & 2.61 \\
    LLaVA-OneVision-72B & 43.03 & 7.51 & 2.96 & 1.78 & 4.55 & 3.8 & 5.94 & 21.67 & 3.48 \\
    \midrule
    Gemma3-4B & 47.10 &  39.32 & 66.54 & 21.05 & 31.91 & 22.73 & 41.21 & 48.15 & 40.40 \\
    Gemma3-27B & 48.73 & 42.98 & 47.23 & 39.17 & 40.00 & 34.18 & 46.58 & 70.00 & 40.00 \\
    \midrule
    Ovis2-8B & 30.56 & 19.77 & 9.76 & 10.63 & 11.76 & 4.17 & 15.90 & 37.93 & 12.69 \\
    Ovis2-34B & 41.74 & 45.72 & 57.51 & 41.60 & 56.36 & 45.57 & 66.21 & 68.33 & 46.96 \\
    \midrule
    \midrule
    \textbf{\texttt{Human Performance}} & 81.23 & 80.11 & 82.31 & 85.12 & 85.37 & 87.23 & 82.17 & 85.78 & 83.21 \\
    \bottomrule
\end{tabular}
\caption{\textbf{Performance of Current LVLMs as PRMs.} We evaluate models using accuracy (ACC) and F1 score as Overall metrics, along with performance on specific error type detection: TE, RU, AE, RR, ODF, ARI, and LU.}
\label{tab:main_results}
\end{table*}

The main results are shown in Table~\ref{tab:main_results}. Several key findings can be summarized as follows:  

\noindent\textbf{LVLMs as PRMs demonstrate limited capability in vision reasoning process evaluation.} Our comprehensive evaluation reveals that current LVLMs struggle significantly when employed as PRMs for visual reasoning tasks. Among open-source models, Ovis2-34B demonstrates relatively strong performance with 41.74\% ACC and 45.72\% F1, slightly outperforming the proprietary Gemini-2.5-Flash model. However, even these best-performing models achieve considerably lower accuracy than human-level performance (83.61\%). This substantial gap suggests that existing LVLMs are not yet competent to serve as reliable PRMs for visual reasoning tasks.

\noindent\textbf{Significant performance disparities exist across different error types.} The results reveal dramatic variations in model performance across evaluation categories. While most models achieve ACC scores below 50\% (ranging from 22-57\%), they fail drastically when identifying specific complex reasoning errors such as TE and RU, where detection rates often drop below 5\%. This indicates that LVLMs struggle particularly with detecting nuanced logical reasoning errors in multi-step processes.

\subsection{Detailed Analysis}

Now we present an in-depth examination of LVLMs serving as PRMs, grounded in our empirical results. Our analysis reveals the following:

\begin{table}[!t]
\centering
\begin{tabular}{lcc}
\toprule
\multirow{3}{*}{\textbf{Models}} & \multicolumn{2}{c}{\textbf{Accuracy}} \\
\cmidrule{2-3}
 & \textbf{Correct} & \textbf{Incorrect} \\
\midrule
Gemma3-27B & 55.83 & 42.17 \\
InternVL2.5-78B & 62.50 & 47.63 \\
LLaVA-OneVision-72B & 63.61 & 56.52 \\
Ovis2-34B & 58.31 & 38.05 \\
Gemini-2.5-Flash & 78.89 & 38.81 \\
\midrule
\texttt{\textbf{Random}} & 50.0 & 50.0 \\
\bottomrule
\end{tabular}
\caption{\textbf{Model Performance Comparison.} Accuracy of different models on correct and incorrect samples.}
\label{tab:model_bias}
\end{table}

\begin{quotebox}
\textbf{\textit{Finding 1.}} LVLMs as PRMs show consistent positive bias in their reward assignments.
\end{quotebox}

Table~\ref{tab:model_bias} demonstrates that both open-source and closed-source LVLMs exhibit significant reward bias during the evaluation process. Notably, the majority of models achieve accuracy rates below random chance (50\%) when identifying incorrect samples, highlighting substantial deficiencies in existing LVLMs' ability to detect reasoning errors.

Furthermore, the comparison between correct and incorrect sample performance reveals a consistent bias pattern across most models, with a pronounced tendency toward positive rewards (i.e., favoring correct classifications). For instance, Gemini-2.5-Flash achieves a remarkable 78.89\% accuracy on correct samples, yet only 38.81\% accuracy on incorrect samples, illustrating this systematic bias.

\begin{quotebox}
\textbf{\textit{Findings 2.}} LVLMs as PRMs show notable step sensitivity and significant performance variations during the multi-step reasoning process.
\end{quotebox}

All LVLMs exhibit severe volatility with accuracy fluctuating dramatically across consecutive steps, often dropping from high values to near-zero performance within one step. This instability is particularly pronounced in correct sample evaluation, where most models struggle to maintain consistent performance above 0.5. Performance disparities between text reasoning steps (even steps) and image extraction steps (odd steps) are evident, with most models achieving higher accuracy during text reasoning while experiencing degradation during image extraction. Gemini 2.5-Flash demonstrates superior stability with smaller fluctuation amplitudes, while Ovis2-34B stands out among open-source LVLMs for more consistent evaluation capabilities, maintaining smaller fluctuation ranges alongside Gemini.

\section{Conclusion}

In this work, we first investigate the problems existing in the \textit{thinking with images} reasoning paradigm and define fine-grained error types. To address a core research question: Whether Process Reward Models Can Help?—We employ guided search experiments, revealing that while current LVLMs can help, they still have significant potential for improvement. Based on this finding, we introduce \ourbench, a benchmark characterized by fine-grained evaluation categories and challenging error types. We construct 1,206 data samples through rigorous human filtering and annotation. \ourbench~serves as a comprehensive testbed for evaluating different LVLMs as PRMs in supervising vision reasoning under the \textit{thinking with images} paradigm. Through comprehensive evaluation of various models, we reveal two key findings: LVLMs as PRMs demonstrate limited capability in evaluating vision reasoning processes, and performance varies dramatically between error types. These findings indicate the substantial gap between current LVLM capabilities and the requirements for reliable process supervision in visual reasoning tasks. We anticipate our work will promote development in LVLM capabilities as PRMs to help the visual reasoning process.

\section{Limitations} Our analysis is based on reasoning trajectories from current \textit{thinking with images} models, which primarily support basic visual operations such as cropping and zooming. More advanced models with richer visual capabilities and sophisticated reasoning tools may exhibit different error patterns and potentially address some of the identified error categories. Future work should extend this taxonomic analysis to incorporate trajectories from more advanced \textit{thinking with images} models.

\section{Acknowledgments}
This work is funded in part by the HKUST Startup Fund (R9911), Theme-based Research Scheme
grant (No.T45-205/21-N) and the InnoHK funding for Hong Kong Generative AI Research and
Development Center, Hong Kong SAR.

\bibliography{aaai2026}

\end{document}